\documentclass[journal]{IEEEtran}

\usepackage{amsmath}
\usepackage[utf8]{inputenc}
\usepackage{graphicx}
\usepackage{cite}
\usepackage{tabularx}

\ifCLASSINFOpdf
\else
\fi

\DeclareUnicodeCharacter{0301}{\'{F'}}
\begin{document}
\hyphenation{op-tical net-works semi-conduc-tor}
%
\title{Real-time information retrieval from Identity cards}
%
%
\author{Niloofar Tavakolian\(^1\), Azadeh~Nazemi\(^2\), Donal Fitzpatrick\(^2\)\\
\(^1\)Department of Computer Science and Engineering and Information Technology, Shiraz University,
\(^2\) School of Computing, Dublin City University

\thanks{ }
}


%



\maketitle


\begin{abstract}
Information is frequently retrieved from valid personal ID cards by the authorised organisation to address different purposes. The successful information retrieval(IR) depends on the accuracy and timing process. A process which necessitates a long time to respond is frustrating for both sides in the exchange of data. This paper aims to propose a series of state- of-the-art methods for the journey of an Identification card(ID) from the scanning or capture phase to the point before Optical character recognition(OCR). The key factors for this proposal are the accuracy and speed of the process during the journey. The experimental results of this research prove that utilising the methods based on deep learning, such as Efficient and Accurate Scene Text(EAST) detector and Deep Neural Network (DNN) for face detection, instead of traditional methods increase the efficiency considerably.

\end{abstract}


%
\IEEEpeerreviewmaketitle
\section{Introduction}

\IEEEPARstart{V}{alid} ID cards which convey reliable and essential information about the cardholder have a vast range in terms of pattern, colour, template and text layout. However, most valid IDs follow the rule of predefined size based on the aspect ratio around 1.58 [1]. passports are a common, legal and internationally accepted proof of identification. The passport identity page includes an area in the bottom of the page which contains all essential fields of information in the specific structure. This area of interest is known as Machine Readable Zone(MRZ) [2]. MRZ data structure extraction and information retrieval is a relatively straightforward process due to its fixed size and location. However, identity cards such as a driver’s license does not have an MRZ. Besides the lamination and texture pattern found in each ID card adds noise, thereby increasing the difficulty of the extraction process. Therefore, applying some preprocessing steps to remove this pattern before data extraction can support speed and accuracy. This study has two phases. The first phase is working around data extraction from a valid ID such as auto-cropping, locating horizontal, facial photo removal and text/non-text segmentation. The second phase attempts to represent an ideal method for data verification for information retrieval. This paper describe the first module. Furthermore, applying techniques such as age recognition, and female/male recognition modules after essential data extraction may be useful for data verification. 
IR results and the text extracted fields of data, images of the frontal facial photo, and cropped image of photo ID are saved in the new record a database. Two fields of images for keeping cropped photo ID and Personal photo are designed for each entry to support verification by the further offline investigation. 
The remainder of this paper includes two sections. Due to the modules multiplicity and diversity in structure and impact to the final result, this study has tried to present the related work, proposed method, and experimental results relevant to each module in an individual subsection. This presentation provides an opportunity for the reader to follow up, evaluate, and conclude each module separately in a logical manner. Section II contains all stated steps as sub-modules, which are started by meaning and concepts description followed by a literature review and proposed a method by this research in terms of time-saving and accuracy. Experimental results of each sub-modules are presented at the end of the relevant step to evaluate results. Section III consists of the conclusion and further development. The following block diagram(Figure 1) illustrates the overview of ID information retrieval. 

Fig. 1: The overview block diagram of information retrieval from an identity passport page 

\begin{figure}[tp]
    \centering
  \includegraphics[scale=0.4]{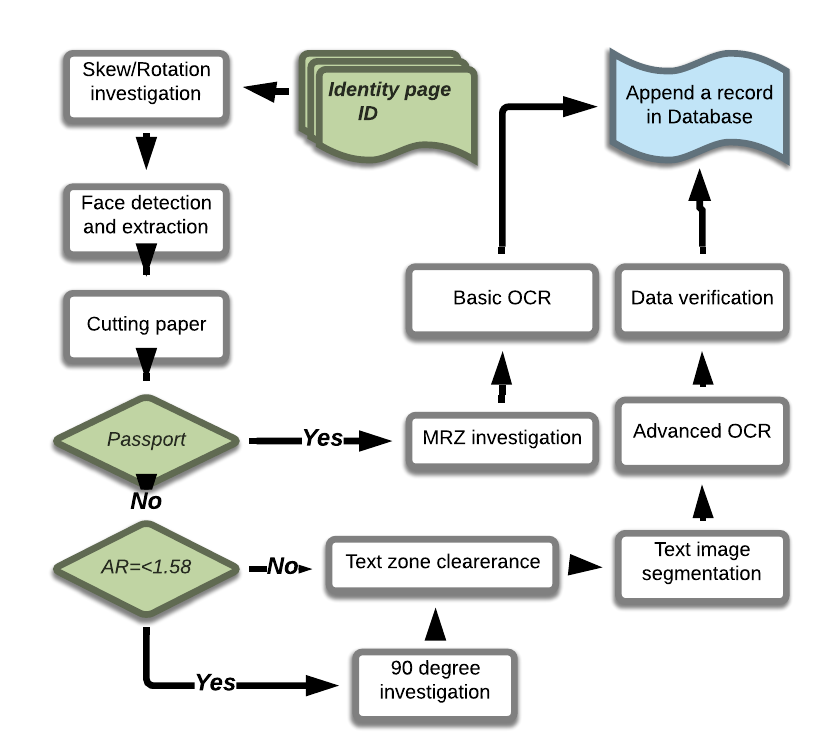}
\caption{The overview block diagram of information retrieval
from an identity passport page}
    \label{fig:galaxy}
\end{figure}
 

\section{Related Work and Proposed Method}

This section presents the processing modules before Optical character recognition(OCR) in the individual subsections. All modules for the journey of capturing device to saving in database are as follows:

\begin{itemize}
  \item Skew and rotation detection/correction; 
  \item Face detection and extraction;
  \item page area detection and calibration; 
  \item pre-processing including pattern removal and increasing contrast;
  \item 90/180 rotated detection and correction; 
  \item MRZ extraction and data retrieval; 
  \item Text image segmentation and field data extraction; 
  \item Running OCR based on language; 
  \item Data verification (applying age detection, female/ male recognition and online signature verification),
  \item Save in database. 
\end{itemize}
    
\subsection{ Skew/Rotation Investigation}

In most cases, the scanned or captured identity page is not horizontal. Rotation or skew has a negative impact on the process of deriving accurate information. Former researchers recommended many approaches to address this issue. The general method to detect text rotation suggested by F ́elix Abecassis [3] relies on finding the text of block, recognising the block angle and rotating the image based on the angle size. There are some approaches depending on text density in the document such as Duda et al. recommended utilising Hough line transform [4] followed by canny edge detector applied on grayscale image to calculate the Hough Transform values and peak determination. Then the detected peaks are categorised into different bins. The bins values denote the angle for correction. F. Weinhaus recommended applying Fast Fourier Transform (FFT) [5] spectrum conversion to documents which include many rows of texts and extracting the list of peaks orthogonal coordinates of text rows to calculate the Mean Square Error (MSE)from the list. In this method, line orientation indicates the angle for correction. However, these methods will not be able to recognise the 90/180 degree rotation in the document. The accuracy of these methods has a straightforward proportional relationship with the text density in the document. In the lack of sufficient text in the document, these methods have failed. This research to address the rotation issue has designed a pipeline includes FFT/MSE, Hough line transform/ Canny edge detection, and finally in F ́elix Abecassis method. The evaluation results of these approaches by this research indicates that F ́elix Abecassis transform has rarely failed and time efficient, which is less accurate than two others. Therefore, designing pipeline contains three methods provide a guarantee to obtain 96\% true positive. Figures 2 and 3 show the result of the deskew processed by the designed pipeline in this research. 
\\
\begin{figure}[htp]
    \centering
  \includegraphics[width=6cm]{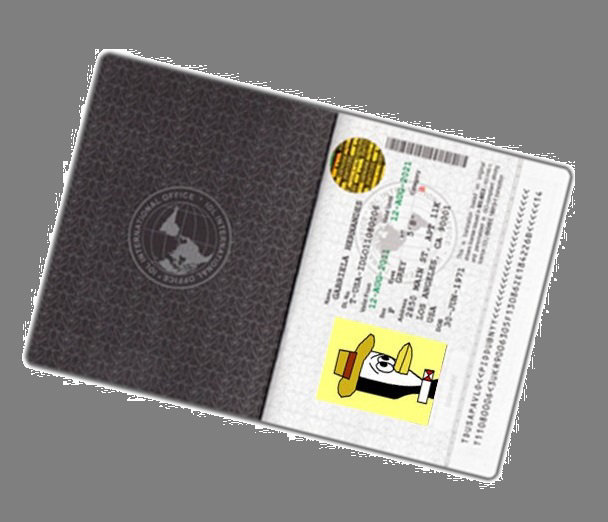}
\caption{+76.3797 degree rotated Identity page}
    \label{fig:galaxy}
\end{figure}

\begin{figure}[htp]
    \centering
  \includegraphics[width=8.5cm]{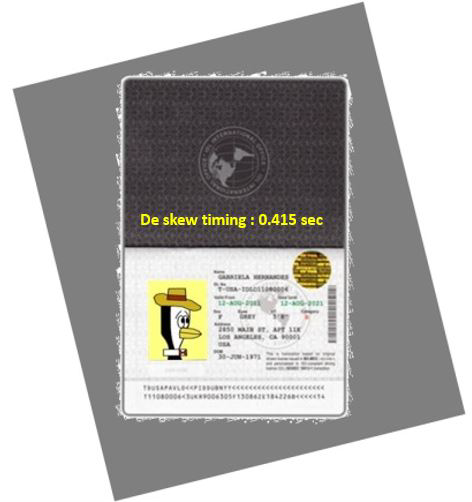}
\caption{After applying first module}
    \label{fig:galaxy}
\end{figure}

\subsection{Photo Face Detection and Extraction }
Face Detection module is performed to extract the facial
photo. Applying Viola-Jones object detection based on HAAR
cascade and Histogram of Oriented Gradient(HOG)/Linear
Binary Pattern(LBP) leads to obtain different results in terms
of accuracy and time(HOG is more preferred). Another
method is Deep Neural Network(DNN) [6] that is a Residual
Network(ResNet-10) Architecture utilising Single-Shot-
Multibox Detector(SSD) [7]. DNN result indicates that DNN
is more robust under unpredictable conditions. Utilizing a
DNN based face detection method indicates that in spite of the time-consuming process of loading a large size model
file(almost ten times bigger than a required XML file for
HAAR), DNN is faster than HAAR cascade and HOG(the
time consumed by DNN is approximately 1
4 HAAR). Due to ethical considerations regarding utilization of frontal-face photographs of real people , this module does not present
the results of applying DNN face detection and extraction
of inappropriate resolution. Since the face boundary box is a
square and the facial photograph is rectangular with an aspect
ratio of 1.33, therefore for extracting the whole area of the
picture, it is required to replace the facial photo with the black
block based on new boundary box as follows:

\begin{equation}
BB_{face} = X_0; Y_0;W;H
\end{equation}

\begin{equation}
H_{PhotoID} = 1.3H
\end{equation}

\begin{equation}
\Delta{H} = 1.3H-H
\end{equation}

\begin{equation}
BB_{PhotoID} =  X_0,Y_0-\frac{\Delta{H}}{2},W,H + \Delta{H}
\end{equation}
 
 The research evaluation for two methods denotes that DNN timing (0.17 seconds) about \(\frac{1}{8}\) HAAR/HOG (1.2 seconds) by
improve accuracy from 89\% to 96.3\%. Figure 4-Top illustrates
applying DNN face detection on a real ID card, and the
Figure 4-Bottom is proof for utilizing Equation 4.

\begin{figure}[tp]
  \centering
  \includegraphics[width=7cm]{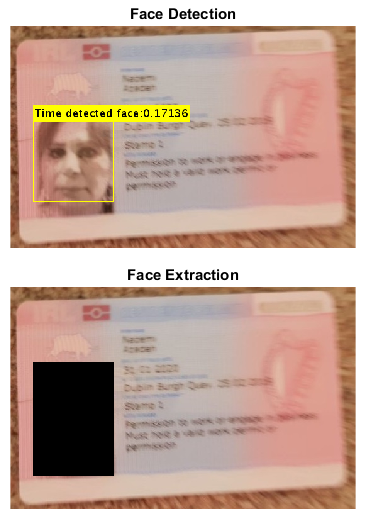}
\caption{Face detection and facial photo extraction using DNN}
    \label{fig:galaxy}
\end{figure}

Fig 3 illustrates the inner class similarity of menders is more than their similarities to other classes (for saliency map).
Fig 4 illustrates How the similarity of members of two different classes  are more than the members inner class similarity(for intensity vectors)  
\subsection{Auto cropping}
The traditional primary method to detect the exact area of
information is based on edge detection or horizontal/vertical
line detection. Edge detection can not support auto cropping
due to uncontrolled illumination conditions or in the case that
locating paper under scanner/camera lens is not reliable [8].
This paper proposes a method based on high detail region
detection. This method initially finds the best region containing
the underlying detail in the image. Standard deviation,
Sobel grayscale edges, or canny binary edges measure these
details [9]. This area has the most critical grayscale mean.
This approach tries to find the sub-image from the original
image with the lowest Root Mean Square Error(RMSE) [10].
Although, if the document contains only two regions based
on detailed density, this method fails. Since this process is
time-consuming, thus it is used for further offline processing,
and another method has been replaced based on contour properties
detection [11]. Besides, the contour properties support
text/non-text segmentation modules. The following images
respectively illustrate high detail extraction, and contour properties
approach results applied for auto cropping. Figure 5-Top
shows the captured image; Figure 5-Middle illustrates high
detailed auto-cropping that is about 30 times slower than using
auto-cropping based on contour property, which is illustrated
in Figure 5-Bottom. The contour property based auto-cropping relies on the trusted area of text. The following steps recognise
this area:
\begin{itemize}
  \item Finding the biggest contour as a connected component.
(level one). 
  \item Finding level two contours inside level one.
  \item Keeping the level two contours which have most same level
contour as a text line.
  \item Keeping Level two contours with aspect ratio \((0:1 < \frac{Width}{Height} < 10) \)as a text character.
\end{itemize}
According to test implementation result for two stated methods,
contour based auto cropping runs in real-time mode by
timing 0.7 seconds. This research use the output of auto-cropping
based on the specific area in the offline mode by
timing 21 seconds for further verification.

\begin{figure}[htp]
    \centering
  \includegraphics[width=6cm]{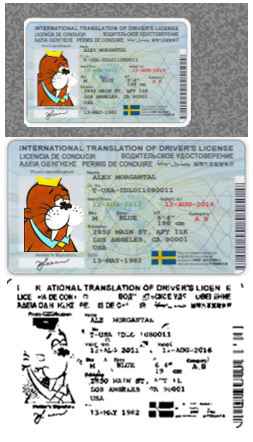}
\caption{Top: Non-cropped captured the image, Middle: the
result of high detail auto-cropping, Bottom: the result of
grayscale contour property based auto-cropping}
    \label{fig:galaxy}
\end{figure}
\subsection{Text cleaning}
According to image processing researchers, noise reduction
is a general approach to enhance text detection result [12].
Auto-cropping based on contour properties partially removes
noise and cleans the text area. Usually, the background of the
ID card has been affected by scanner/camera lens dust or dirt.
This affection causes many conflicts with the result of the
next module. Therefore, this research proposes text cleaning
after auto-cropping for further enhancement. This text cleaning
process is applied to the ID card and starts by converting
the cropped image to the grayscale. The procedure continues
by normalizing for enhancement purposes and converting to
grayscale again. Afterwards, the enhanced grayscale image is
converted to binary using the local threshold to create a mask
by blurring, and finally, the obtained mask convolves to the
original image to generate a clean background. In this stage,
the result may require sharpening. Figure 6 shows two steps
of background cleaning to improve the OCR result.
\begin{figure}[tp]
    \centering
  \includegraphics[width=7cm]{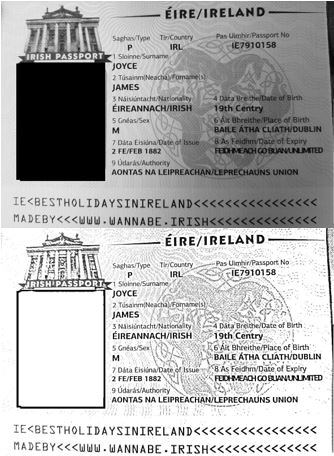}
\caption{Background cleaning Double steps.}
    \label{fig:galaxy}
\end{figure}

\subsection{Text/non-text segmentation}
One of the most significant modules that affect the final
results is text/non-text segmentation, which is the area of
research by many computer vision researchers. There are many
successful approaches for this purpose, such as Maximally
Stable External Region (MSER) [15], Stroke Width Transform(
SWT) [14], and the most recent one is Efficient and Accurate
Scene Text Detector(EAST) [13] based on deep learning.
EAST detect words or text lines of arbitrary orientations
and quadrilateral shapes, without need word, segmentation
only using one Neural Network(NN). According to Huizhong
Chen et. al. MSER represents robust and enhanced approach
to text detection.Chen used MSER with Canny edge to address
MSER drawback caused by blurring. EAST is a state-of-theart
deep learning model based on an innovative architecture
and training pattern with timing performance about 0.07 for
each frame. EAST is a suitable algorithm for extracting
natural scene images. According to Celine Mancas-Thillou
et al. 2017 in contrast, to text segmentation in laboratory
conditions, this process from natural scene images is challenging
due to non-predictable conditions such as low-resolution
device capturing, noise, illumination, occlusion, deforming
and positioning. However, in laboratory conditions, heuristic-based
methods leading to obtaining a reasonable result for the
real-time or natural scene is not straightforward. This study
compares the result of MSER and EAST in terms of timing and
accuracy. However, this comparison did not conclude with any
exact result regarding the accuracy. The experimental results
show that the EAST takes less time than MSER. SWT initially
finds high contrast edges. Then traverses the image pixels’
edge, in the normal direction to obtain another normal edge.
This method leads to identify strokes, which is an element
of finite width with two roughly parallel sides. These strokes
play roles as a text area. SWT, due to its time-consuming
nature, is not an efficient solution for this module. Besides,
this module considers contour property based detection (used
for auto-cropping) as an alternative. Figure 7 shows t based on contour. he text detection and character segmentation
contrast to two stated approaches contour-based segmentation
is character-based.
\begin{figure}[b]
    \centering
  \includegraphics[scale=0.3]{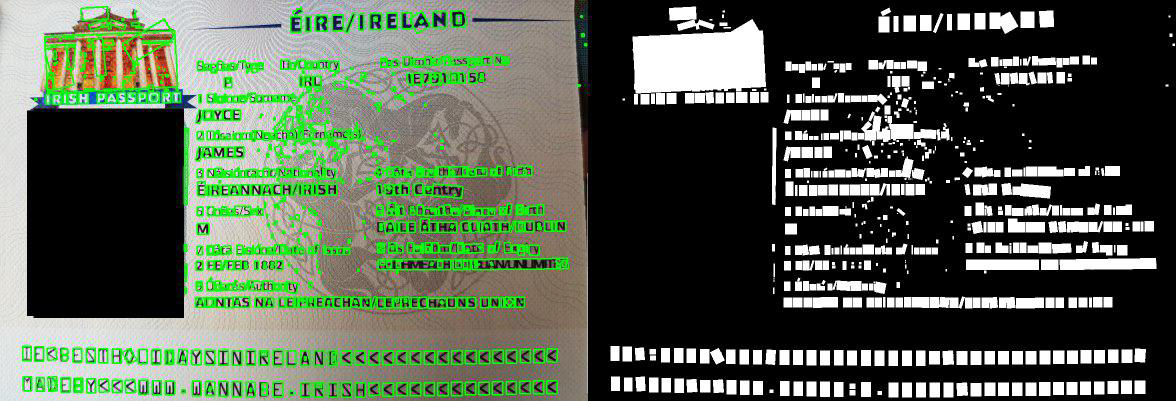}
\caption{characterized contour based text segmentation}
    \label{fig:galaxy}
\end{figure}
    
 Figure 8 illustrates a sample of Text Detection of a passport page based on MSER.
 
 \begin{figure}[b]
    \centering
  \includegraphics[width=9.4cm]{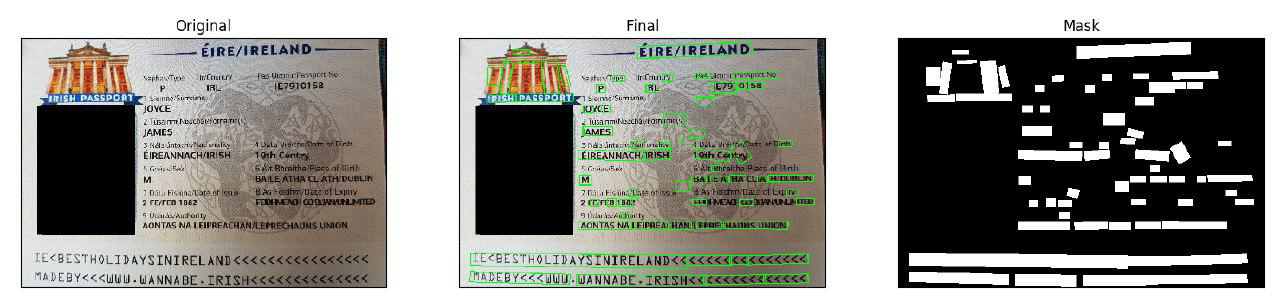}
\caption{MSER text segmentation}
    \label{fig:galaxy}
\end{figure}

 Figure 9 illustrates a sample of Text Detection of a passport page using Deep Learning based on EAST.
 
 \begin{figure}[ht]
    \centering
  \includegraphics[scale=0.2]{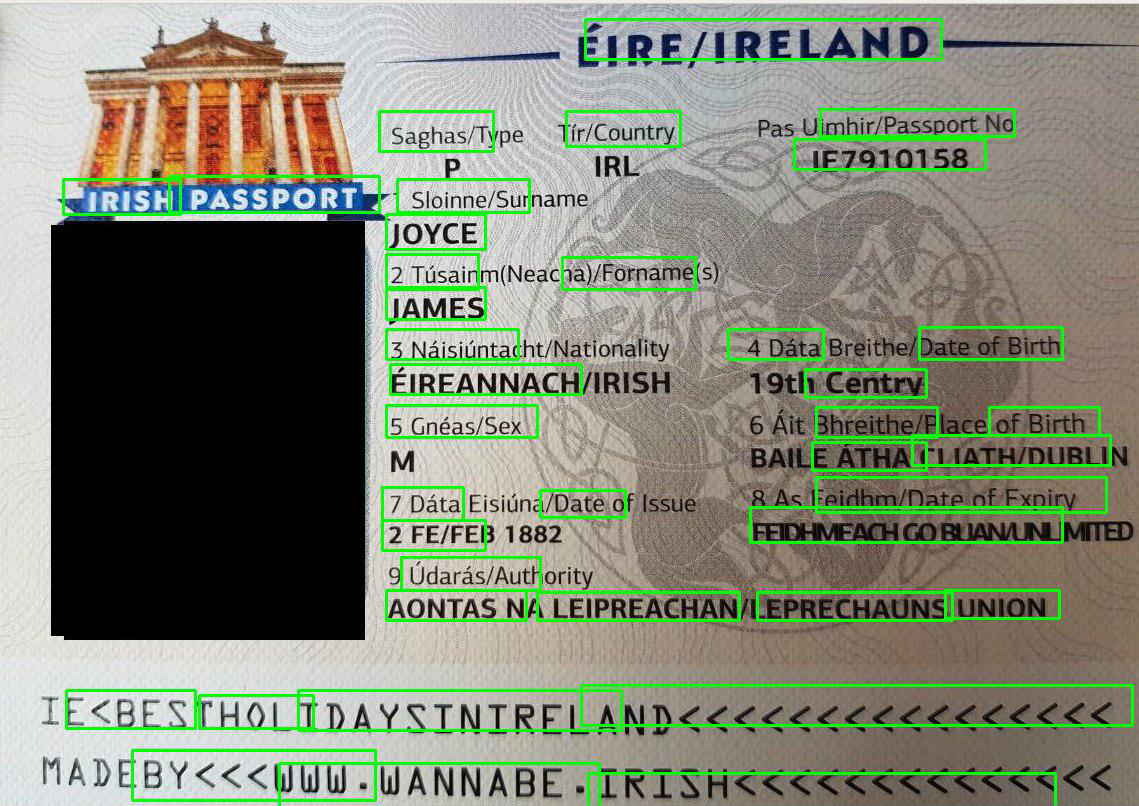}
\caption{EAST text segmentation}
    \label{fig:galaxy}
    
\end{figure}
\subsection{ Detecting 90 degree rotation}
In general cases, two error conditions may indicate that the
ID card does not locate horizontally.

\begin{itemize}

  \item Before auto cropping:
        The face detection module, even if the image has been
rotated180 degrees, detects the facial photo. Additionally,
before the face detection module, any other rotation had
been investigated. Therefore, in such a case which this
module does not return any boundary box, it means the
ID image was located vertically during image capture.

  \item After auto cropping:
        If the calculated aspect ratio of the cropping ID is not
equal to the pre-defined layout.

\end{itemize}

In these cases to prevent failure in subsequent modules, it is
required to rotate by 90 degrees before further investigation.
Since 90 degree rotation is not detectable by the former
modules, after auto cropping aspect ratio( Width
Height ) of the ID
card is calculated and if it is less than pre-defined layout 1.58,
it is required to rotate the image 90 degrees. Besides, rotation
of 90 degrees of the ID card before auto cropping leads to
failure of the performance of face detection.

\subsubsection{ MRZ investigation}
  
Based on the International Electrotechnical Commission
MRZ includes two strings of 44 characters. The importance of
MRZ is recognizable and readable by a machine, regardless
of passport nationality. MRZ has two categories: Category one
has three lines, and class 2 has two. MRZ provides information
about the type of passport, passport type, validation date,
passport holder name, full name, citizenship, and gender of
the passport holder. Therefore, MRZ extraction is a verification
tool for further data extraction. This research has used some
morph techniques to detect and extract MRZ based on [16].
Blackhat morphological operation is suitable for releasing
the black MRZ text against the light passport background.
After applying this morph operation, the magnitude of the
result is calculated. This module uses two kernels for further
morphological operations. The first rectangular kernel with
aspect ratio more than two removes blank areas between MRZ
characters, and the square Kernel is used to insert blank space
between MRZ lines.
Figure 6-Top illustrates the mask of passport identity page
after applying morphology techniques. Figure 6-Bottom shows
extracted MRZ area.
 
\begin{figure}[ht]
    \centering
  \includegraphics[width=8cm]{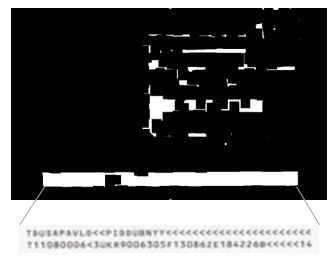}
\caption{ Figure 6-Top: The mask of passport identity page after
morphology techniques, Figure 6-Bottom: MRZ extracted area}
\label{fig:galaxy}
\end{figure}

\subsection{CONCLUSION}
This research has been conducted to improve timing and
accuracy information retrieval from identity card or passport.
It aims to reduce the waiting time for checking and verification
customer based on his/her claim upon presenting the ID. Based
on this study, The journey of an ID card from capturing
device to OCR includes many steps. Timing, accuracy, and
performance of each step has been evaluating separately Since
the IR based on OCR result has straight proportional relation
with the precision of text segmentation, this study attempts to
find the most appropriate solution for the text segmentation.
Although EAST is the robust and accurate text segmentation
method for the wild scene, still it is sensitive to rotation,
noise, and not proper margin. Therefore, this study attempts
to find the most effective solution considering the timing for
these issues. Experimental results prove that replacing MSER
by EAST improves the timing from 0.7 to 0.07. However,
the voting system finds the most interested area by MSER,
EAST, and contour property. Comparing two auto-cropping
approaches indicates that timing ratio is 30. It means, regardless
of high obtained precision of finding the most detailed
area that extracts the actual corners, this process is time consuming.
Conclusively, finding the most significant connected
component followed by detection of contour with the
specific attribute plays an auto-cropping role. Regarding face
detection and extraction, utilizing DNN increases accuracy
about 10\% and reduce the timing by about 0.75\%. Although,
for rotation and skew detection, FFT followed by MSE, for the
document full of text record leads to 96\% true-positive results,
and using Hough transform followed by Canny edge obtain
the more reliable robust report for the rotation probability.
Therefore, to prevent failing this research uses a pipeline of
three methods to address skew issues. Finally this research
concludes that classifies implemented methods to real-time and
offline. Further development and future work will follow this
study by evaluating verification methods such as age, gender
and signature recognition.
\ifCLASSOPTIONcaptionsoff
  \newpage
\fi

%


%


\end{document}